%% file: my-paper.tex
\documentclass[pdflatex,sn-mathphys-num]{sn-jnl}

\usepackage{graphicx}%
\usepackage{multirow}%
\usepackage{amsmath,amssymb,amsfonts}%
\usepackage{amsthm}%
\usepackage{mathrsfs}%
\usepackage[title]{appendix}%
\usepackage{xcolor}%
\usepackage{textcomp}%
\usepackage{manyfoot}%
\usepackage{booktabs}%
\usepackage{algorithm}%
\usepackage{algorithmicx}%
\usepackage{algpseudocode}%
\usepackage{listings}%

\theoremstyle{thmstyleone}%
\theoremstyle{thmstyletwo}%

\theoremstyle{thmstylethree}%

\begin{document}

\title[Article Title]{Maternal and Fetal Health Status Assessment by Using Machine Learning on Optical 3D Body Scans}

\author*[1]{Ruting Cheng}\email{rcheng77@gwu.edu}

\author[1]{Yijiang Zheng}\email{yijiangzheng@gwu.edu}

\author[1]{Boyuan Feng}\email{fby@gwu.edu}

\author[1]{Chuhui Qiu}\email{chqiu@gwu.edu}

\author[2]{Zhuoxin Long}\email{zlong66@gwu.edu}

\author[3]{Joaquin A. Calderon}\email{joacocal91@gmail.com}

\author[2]{Xiaoke Zhang}\email{xkzhang@gwu.edu}

\author[3]{Jaclyn M. Phillips}\email{japhillips@mfa.gwu.edu}

\author[1]{James K. Hahn}\email{hahn@gwu.edu}

\affil*[1]{\orgdiv{Department of Computer Science}, \orgname{The George Washington University}, \orgaddress{\street{800 22nd Street NW}, \city{Washington DC}, \postcode{20052}, \state{}, \country{USA}}}

\affil[2]{\orgdiv{Department of Statistics}, \orgname{The George Washington University}, \orgaddress{\street{801 22nd Street NW}, \city{Washington DC}, \postcode{20052}, \state{}, \country{USA}}}

\affil[3]{\orgdiv{Department of Obstetrics and Gynecology}, \orgname{The George Washington University}, \orgaddress{\street{2150 Pennsylvania Ave. NW}, \city{Washington DC}, \postcode{20037}, \state{}, \country{USA}}}

\abstract{Monitoring maternal and fetal health during pregnancy is crucial for preventing adverse outcomes. While tests such as ultrasound scans offer high accuracy, they can be costly and inconvenient. Telehealth solutions and more accessible body shape information provide pregnant women with a convenient way to monitor their health. This study explores the potential of 3D body scan data, captured during the 18-24 gestational weeks, to predict adverse pregnancy outcomes and estimate clinical parameters. We developed a novel algorithm with two parallel streams which are used for extract body shape features: one for supervised learning to extract sequential abdominal level circumference information, and the other for unsupervised learning to extract global shape descriptors, alongside a branch incorporating shape-related demographic data. Our results demonstrated that 3D body shapes can support the prediction of preterm labor and gestational diabetes mellitus (GDM), as well as the estimation of fetal weight. Compared to other machine learning models, our algorithm achieved the best performance, with prediction accuracies exceeding 89\% and fetal weight estimation accuracy of 72.22\% within a 10\% error margin, outperforming the conventional anthropometric measurements-based method by 18.18\%.
}


\maketitle

\section{Introduction}\label{sec1}

Systematic prenatal care has been recognized as essential for reducing maternal and neonatal morbidity and mortality rates since the early twentieth century \cite{thompson1990history,peahl2021evolution}. However, some studies concluded that many in-person prenatal care visits are unnecessary \cite{peahl2020prenatal,reisman2020covid}, and the prenatal care regimen has seen limited improvements despite decades of advances in diagnostic and communication technologies \cite{peahl2021evolution,shmerling2022prenatal}. Recent studies have suggested that incorporating telehealth modalities can improve prenatal care by reducing unnecessary trips to hospital and interventions \cite{whittington2020telemedicine,ferrara2020telehealth,hantsoo2018mobile,tobah2019randomized}. For pregnant women in rural communities, prenatal care through telehealth is even more crucial, as it helps overcome barriers to accessing equitable healthcare resources \cite{whittington2020telemedicine}.

Although telehealth offers convenience, it has disadvantage of lacking in laboratory tests \cite{breton2021telehealth,gajarawala2021telehealth}. To address this issue, it is crucial to explore biomarkers which can be easily obtained through widely-available devices for health monitoring. The 3D body shape is an ideal data modality that can be collected in non-invasive ways, especially in the field of obstetrics, where distinct body shape changes occur on pregnant women. Meanwhile, previous studies have shown a strong correlation between maternal anthropometric measurements and the health status of mothers and fetuses \cite{pschera1984estimation,tabrizi2012maternal,dare1990value}. Ay et al. found that maternal body mass index (BMI) during pregnancy positively correlates with the fetal weight. Maternal height, pre-pregnancy BMI, and gestational weight gain are also found to correlate with potential risks of having a small or large for gestational age child \cite{ay2009maternal}. Risk of preeclampsia (PEC) and cesarean delivery have been associated with maternal weight, height and body circumferences \cite{derbyshire2009can, boucher2022maternal}. These promising results indicate the potential for using body shape features as additional resources in telehealth for pregnancy. Moreover, comparing to traditional anthropometric features, detailed 3D body scan data can provide more body shape information, such as the shape, volume and position of gravid uterus. 

The development of 3D optical scanning technology has made it possible to capture 3D body shape data using commercial 3D body scanners \cite{ng2016clinical}. Besides, some commodity smartphone apps allow users to accurately scan themselves at home by using built-in LiDAR or camera with 3D reconstruction algorithms \cite{smith2022anthropometric}. With the accessibility and accuracy of 3D body models, there is an increased interest in developing algorithms capable of efficiently extracting information from detailed body shapes for obstetric analysis. 

Currently, 3D body scanning technology in the field of obstetrics and gynecology is still in its early stages. Rather than being employed as a novel data modality for direct diagnosis, 3D scan data have primarilu been used as visualization tool or a medium to extract conventional anthropometric features. For example, Dathan-Stumpf et al. used 3D body scans to obtain pelvic measurements for vaginal breech delivery assessment \cite{dathan2023novel}, while Gradl applied structure light to monitor maternal abdomen changes during pregnancy and quantified abdominal deformation to evaluate fetal kicking activity \cite{gradl2022application}. However, these applications do not fully exploit the rich information contained in 3D body scan. In contrast, studies in the field of nutrition have shown strong potential of applying machine learning algorithms for 3D body shapes analysis. For instance, Wang et al. designed a bi-channel network to detect hepatic steatosis using 3D body shape reconstructed from CT iso-surfaces \cite{wang2022s2flnet}. Feng et al. proposed a hierarchical neural network to estimate body compositions including lean mass, fat mass, and bone mineral content using whole-body optical scans\cite{fengenhanced}. These studies highlight the feasibility of applying machine learning–based 3D body shape analysis to other medical domains, including obstetrics.

In this study, we aim to leverage machine learning–based 3D body shape analysis for maternal and fetal health status assessment. Our main contributions are summarized as follows:

\begin{enumerate}
\itemsep=0pt
\item We used 3D body scans and basic body shape-related demographic information to predict the risk of preterm labor, gestational diabetes mellitus (GDM), PEC, and the likelihood of delivery by cesarean section. We also used the same body shape features to estimate fetal weight and maximum vertical pocket (MVP). To the best of our knowledge, this is the first study that uses 3D body scans of pregnant women to assess maternal and fetal health status. 
\item We designed a novel hybrid neural network comprising a supervised learning stream and an unsupervised learning stream to extract shape features from 3D shape data, which is robust even on imbalanced small-sized data. 
\item We conducted experiments comparing our algorithm using 3D body scans as inputs with the baseline method using Logistic Regression (LoR, for classification tasks), Linear Regression (LR, for regression tasks), and anthropomorphic measurements. Additionally, we also compared our algorithm with other well-performing machine learning algorithms using the same 3D body shape inputs. The experimental results not only indicate the feasibility of using 3D body scans to assess maternal and fetal health status, but also demonstrate the precision and efficiency of our new approach.
\end{enumerate} 

\section{Related work}\label{sec2}
\subsection{Association between anthropometric measurement and pregnancy health status}\label{subsec2}

As the foundation for applying 3D body shape analysis in obstetrics, the relationships between maternal anthropometric measurements and various adverse outcomes have been extensively studied, with strong correlations identified. Boucher et al. conducted a study which revealed associations between the probability of cesarean delivery and anthropometric measurements such as weight, BMI, waist circumference and skinfold thickness \cite{boucher2022maternal}. Sina et al. found that increasing BMI, weight, waist circumference, and waist-to-height ratio were associated with an increased risk of GDM, with BMI and waist-to-height ratio showing stronger correlations \cite{sina2015associations}. Ebrahimi-Mameghani et al. concluded that early pregnancy BMI and waist circumference were associated with the risk of gestational hypertension (GH), PEC, and preterm labor \cite{ebrahimi2013correlation}.

Besides investigation of the relationship between maternal anthropometry and pregnancy outcomes, some studies also showed the associations between maternal body shape and intrauterine parameters. In the absence of ultrasound equipment, clinicians can estimate fetal weight by using formula such as Johnson's formula, Insler and Bernstein's formula for a series of clinical maternal measurements such as symphysis-fundal height (SFH) and abdominal circumference (AC) \cite{lanowski2017ultrasound,shittu2007clinical,dare1990value,johnson1954estimation,sherman1998comparison,khani2011comparison}. Anggraini et al. proposed a model to estimate fetal weight based solely on fundal height (FH) \cite{anggraini2020development}. Given single value as input, this method is relatively effective in detecting abnormal fetal growth, however, this model is still constrained by limited amount of information and the requirement for manual measurement. Recently, another study used deep neural networks to estimate fetal weights with multiple variables as inputs, including blood laboratory tests and medical history. The analysis result shows that pre-pregnancy weight and BMI have great impact on estimated fetal weight (EFW) \cite{wang2023fetal}.

These collective findings underscore the potential of utilizing body shape features to predict the likelihood of adverse pregnancy
outcomes and estimate intrauterine parameters such as fetal weight, which can benefit applications in telehealth.

\subsection{Applications of 3D body shape in obstetrics field and advanced methodologies for extracting information from 3D body shape}\label{subsec2}

Anthropometric techniques have been widely used to capture simplified body shape features for a long time, but these methods cannot fully represent the whole body shape \cite{utkualp2015anthropometric,young1979height}. Moreover, although these methods usually do not require complex equipment, they still need to be carried out by professionals.

Thus, some researchers investigated the applications of 3D body scanning in obstetrics. Gradl conducted a preliminary study monitoring maternal abdomen changes up to two months postpartum and evaluated fetal kicking activity during late pregnancy \cite{gradl2022application}. Glinkowski et al. used 3D surface topography method to reveal postural changes of pregnant women and studied the relationship between spinal curvatures and low back pain during pregnancy \cite{glinkowski2016posture}. Dathan-Stumpf et al. captured anthropometric parameters from 3D scan models of pregnant women to predict successful vaginal breech deliveries. The results showed that the prediction accuracy of the method using 3D scanning measurements is at least as good as Magnetic Resonance Imaging (MRI) diagnosis \cite{dathan2023novel}. These studies further indicated the usability of 3D body scans in obstetrics. However, these approaches remain limited to visualization purposes or the extraction of a small set of anthropometric measurements from 3D models, rather than directly leveraging the 3D models themselves as a new imaging modality.

To fully leverage information from the complicated representations of 3D models, which are often comprised of thousands or millions of polygons or 3D points \cite{haleem20193d}, researchers are investigating more efficient body representations and methodologies and evaluating them on different medical applications. Xie et al. proposed to use frontal whole-body silhouettes for estimating body composition. Test results on estimating fat mass index and fat-free mass index supported the usability of this silhouette-based method \cite{xie2015accurate}. Ng et al. proposed a method to estimate regional body composition by using regional circumferences, areas, and volumes obtained from 3D body models with a linear model \cite{ng2016clinical}. Lu et al. used level circumferences to represent 3D body shapes. With the extracted level circumferences, they employed a Bayesian network to predict pixel-level body composition and body fat percentage for the 2D projection of 3D body \cite{lu20193d}. Wang et al. constructed a body shape descriptor called Shape Map from body shape contour of CT scan slices and used it to predict whole-body fat percentage and visceral fat percentage \cite{wang2021pixel}.  Roy et al. obtained the trajectory of the vertebral column from body shape and derived potential lateral deviations of the spine and rotation of the vertebrae. This method performed well in the estimations of the lateral deviation of the spine for mild and moderate scoliosis \cite{roy2019noninvasive}. Su et al. explored the diagnosis of type II diabetes using features selected from 3D scan data through machine learning algorithms like backpropagation neural networks and decision trees \cite{su2006data}. Zheng et al. investigated the usability of raw 3D point cloud of body scans for deep learning-based body fat percentage prediction \cite{zheng2024d3bt}. The same group also conducted a study comparing the effectiveness of different body shape representations, including abdominal level circumferences, area, and trunk 3D point cloud, in nonalcoholic fatty liver disease assessment \cite{zheng2024predicting}. Although these studies are not specifically conducted in obstetrics field, the various methods explored to represent 3D body shape and extract body shape features are worth learning.

\section{Dataset and pre-process}\label{sec3}

\subsection{Dataset}\label{subsec3}

Our dataset comprises a total of 144 study participants, who were recruited at the George Washington University Medical Faculty Associates. The demographic characteristics of this study population is listed in \autoref{tbl1}. To mitigate the influence of confounding factors, we implemented the recruitment following these exclusion criteria: (1) being under the age of 18; (2) having carried multiple gestations; (3) having been diagnosed with an enlarged fibroid uterus; (4) having a BMI exceeding 60; (5) having any unstable medical or emotional condition or chronic disease that would preclude study participation; (6) having undergone body shape altering operation such as liposuction or plastic surgery. This data collection was approved by the Institutional Review Board (IRB). Informed consent forms were collected from all participants.

\begin{table}[h]
\caption{Demographic characteristics of study population.}\label{tbl1}
\begin{tabular*}{0.85\textwidth}{@{}l @{\extracolsep{\fill}} ll @{}}
\toprule
Demographics & N=144 \\
\midrule
Age(yr) & $32.98\pm5.51$ \\
Race &   \\
\quad white & 76 \\
\quad African American & 53 \\
\quad others & 15 \\
Height(m) & $1.63\pm0.07$ \\
Weight(kg) & $72.13\pm13.92$ \\
Gestational age(d) & $146\pm7.92$ \\
\bottomrule
\end{tabular*}
\footnotetext{Notes: For terms with continuous values, we provide mean$\pm$SD; for terms with discrete values, we provide number of participants.}
\end{table}

Each participant underwent optical 3D body scans between the 18th and 24th weeks of pregnancy with Fit3D optical scanner (Fit3D, San Francisco, CA). The 3D scanning process was repeated two to three times to ensure the reliability of the collected data. The participants were required to wear fitting clothes to ensure accurate representation of their body shapes. To investigate the usability of smartphone scans, we also collected 3D body scans using smartphone with the application Polycam (Polycam Inc, San Francisco, CA), as shown in \autoref{FIG:1} below. However, the precision of these smartphone scanned model needs to be tested by additional experiments. Since this study only discusses the theoretical usability of three-dimensional models for obstetric predictions and estimations, we have currently chosen this commercial Fit3D optical scanner with whose precision tested in other studies \cite{sobhiyeh2021digital,ng2016clinical}. The results will be used to determine the feasibility of models obtained through smartphone scanning for future telehealth applications.

\begin{table}[h]
\caption{Clinical parameters of study population}\label{tbl2}
\begin{tabular*}{0.85\textwidth}{@{}l @{\extracolsep{\fill}} ll @{}}
\toprule
Target &   \\
\midrule
EFW(g) & $382.48\pm67.90$ (N=120) \\
MVP(cm) & $4.75\pm0.95$ (N=112) \\
Length of gestation(d) & $271.40\pm15.83$ (N=131) \\
Participants with cesarean section delivery & 34 (N=129) \\
Participants with preterm labor & 15 (N=131) \\
Participants with GDM & 15 (N=138) \\
Participants with PEC & 13 (N=130) \\
\bottomrule
\end{tabular*}
\footnotetext{Notes: $N$ denotes the number of participants who have corresponding medical records. Since we have 13 participants transferred to other hospitals, 2 not delivered yet, 2 underwent cesarean section at their own request, and some records were missing, the sample size $N$ varies across different tasks. For terms including EFW, MVP and length of gestation, we provide mean$\pm$SD; for terms with binary results, we provide number of positive participants.}
\end{table}

In addition to 3D scan data, we collected gestational age (GA) as a temporal marker, along with basic body shape-related demographic information including height and weight. To explore which adverse pregnancy outcomes and parameters can be predicted by body shape features, we also collected the following information: EFW (obtained from ultrasound examination report using Hadlock's formula), MVP, indications of GDM and PEC, length of gestation, and delivery type \cite{hadlock1985estimation}. The detail of these clinical parameters are listed in \autoref{tbl2}. Anthropometric measurements used as inputs for our baseline models were automatically generated by the Fit3D scanner. For each task, subjects who had incomplete information were excluded from subsequent analysis. \\

\begin{figure*}[h]
	\centering
	\includegraphics[width=.8\textwidth]{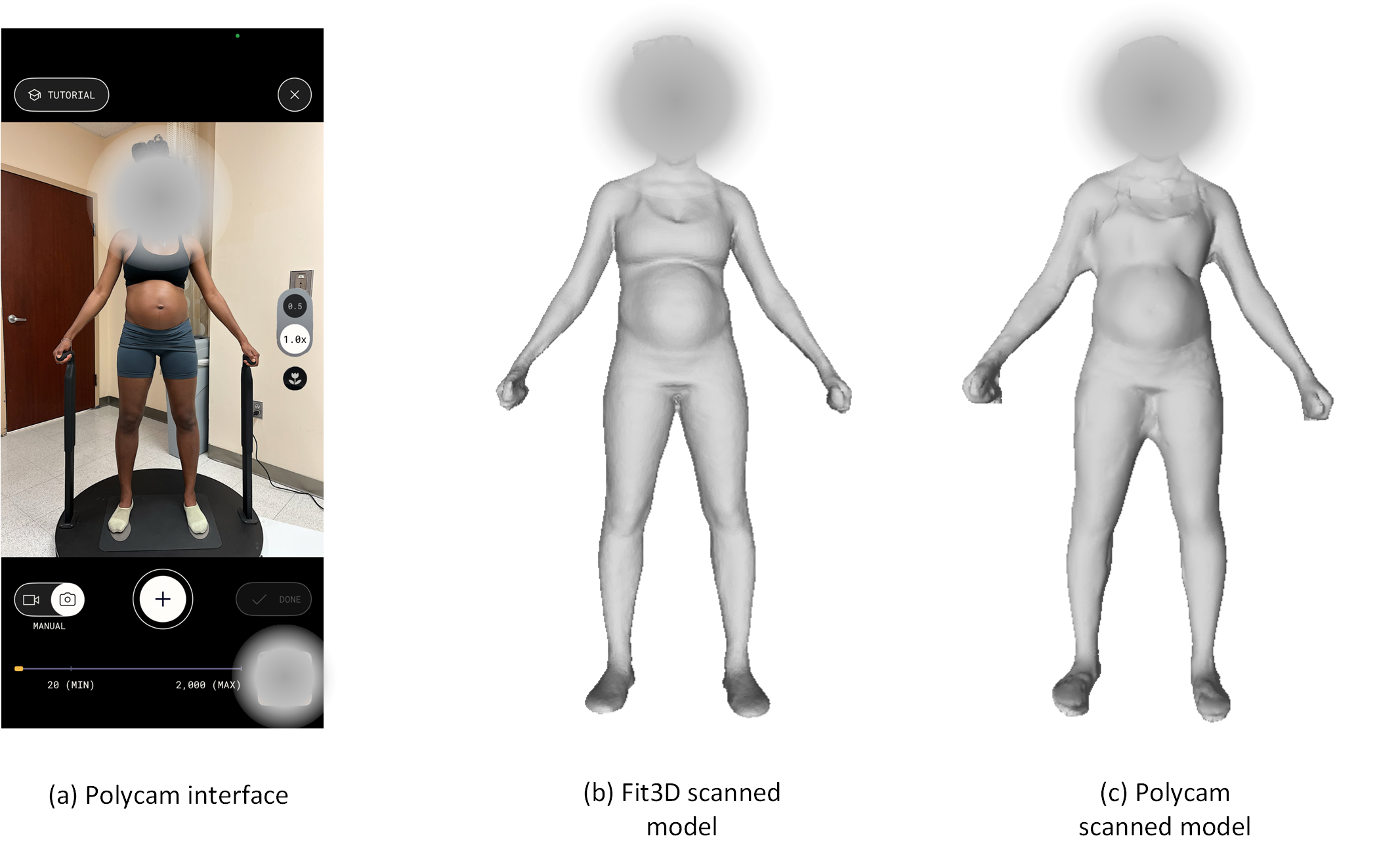}
	\caption{Comparison of Fit3D scanned model and Polycam scanned model captured simultaneously.}
	\label{FIG:1}
\end{figure*}

\subsection{Pre-process}\label{subsec3}

\begin{figure*}[h]
	\centering
	\includegraphics[width=.5\textwidth]{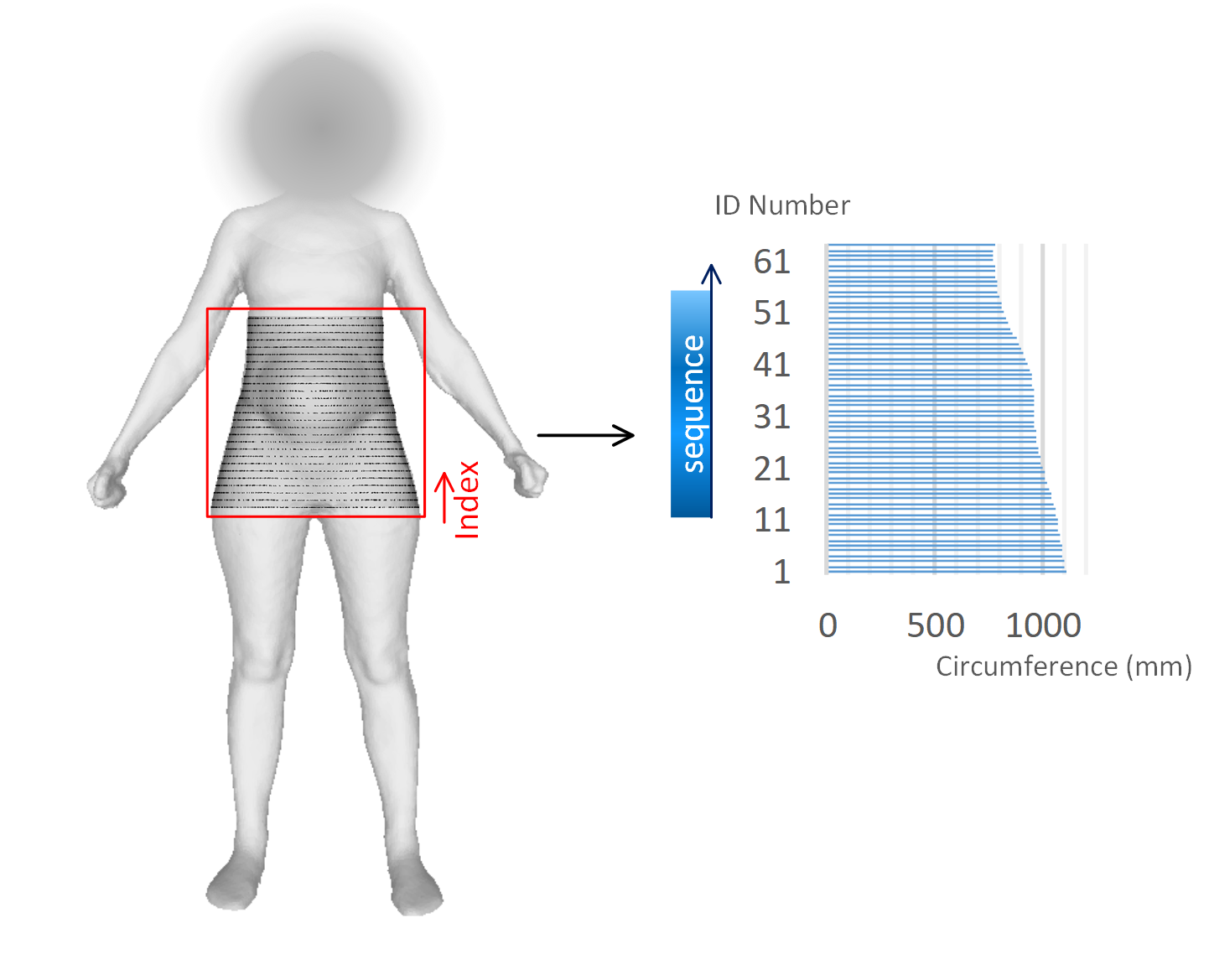}
	\caption{Extracting abdominal level circumference sequence.}
	\label{FIG:2}
\end{figure*}

Based on the anthropomorphic variables from our previous work \cite{sina2015associations,ebrahimi2013correlation,lanowski2017ultrasound,shittu2007clinical,dare1990value,johnson1954estimation,sherman1998comparison,khani2011comparison}, we have identified weight, height, waist circumference, and hip circumference as having potentially strong correlations with maternal and fetal health status. Therefore, after we extracted heights and weights from participant’s medical records, we focused primarily on the abdominal body shape in the region from the lowest point of the pubic bone to the bottom of the breasts, as shown in \autoref{FIG:2}. For each 3D model, we manually labeled the lower and upper boundaries of this abdominal region. Following this, we performed a uniform sampling of 64 level circumferences within this region to create a sequence of the 64 measurements. This approach allows for a simplified yet relatively more informative representation of the 3D body shape by emphasizing the critical region pertaining to pregnancy health status.

\section{Method}\label{sec4}

\begin{figure*}[b]
	\centering
	\includegraphics[width=.9\textwidth]{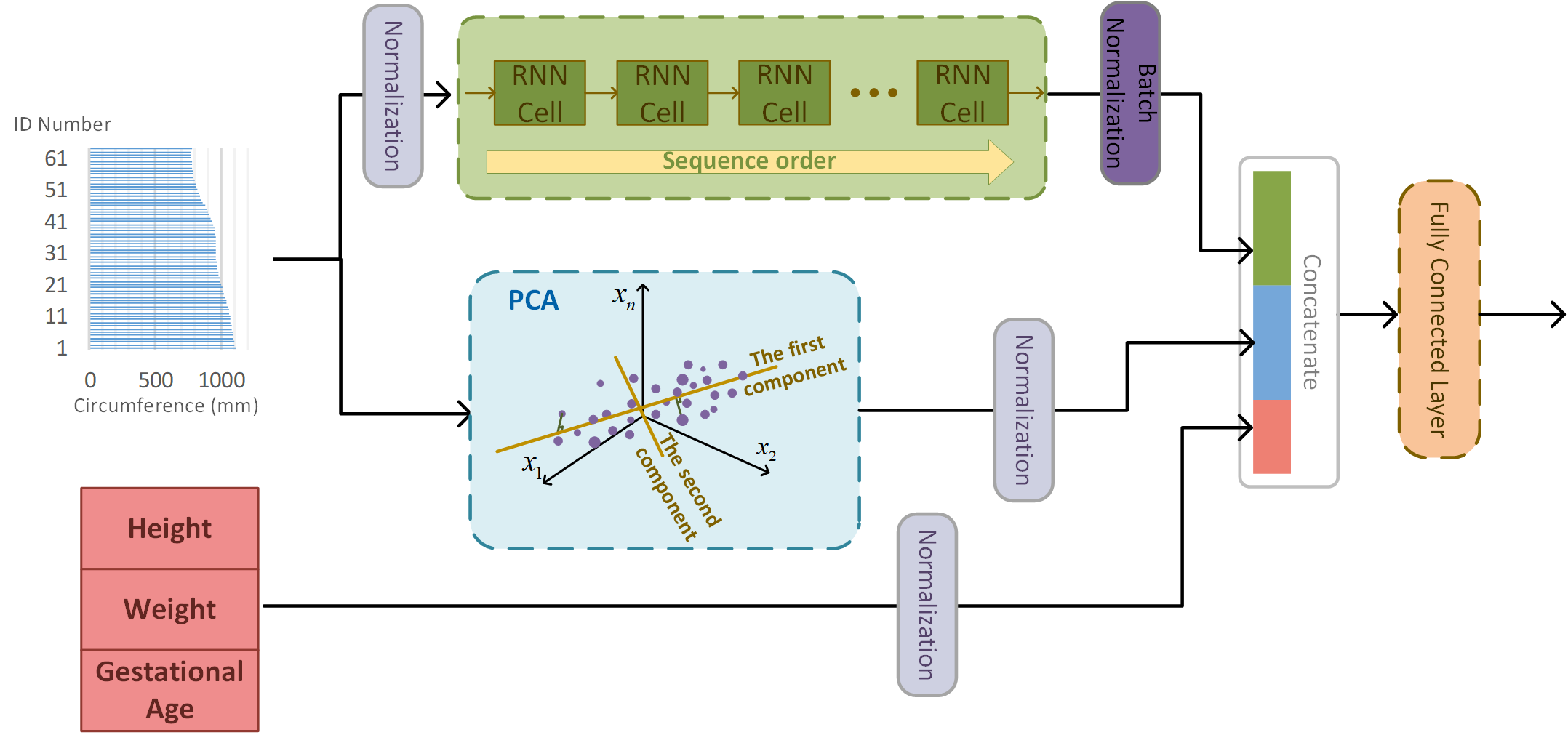}
	\caption{Architecture of the proposed algorithm: Here, the 64 level circumferences and basic demographic information are used as input for the network. The two-stream body shape analysis branch, which is designed to process the level circumferences, consists of a supervised RNN and an unsupervised PCA.}
	\label{FIG:3}
\end{figure*}

Our algorithm is designed to leverage both the maternal body shape and basic demographic information for assessing maternal and fetal health status. Therefore, this algorithm employs a dual-branch structure that processes the two modalities of data separately, as depicted in \autoref{FIG:3}. The first branch is used to extract body shape features from 3D scan data, while the other branch is used to process the demographic features. In the first branch, we designed two parallel processing streams to comprehensively learn the 3D shape information from different perspectives and extract complementary information. The first stream of this branch is used to learn spatial dependency of the abdominal level circumference through a supervised learning algorithm, Recurrent Neural Network (RNN), which is sensitive to the local pattern between neighboring elements. The second stream is used to extract the most representative global features of the same region via an unsupervised Principal Component Analysis (PCA) algorithm. This algorithm has been successfully used in other studies based on 3D body shape analysis, which can guarantee the basic performance of the algorithm and enhance the robustness given a small dataset \cite{tian2022device,lu20193d,zheng2024predicting}. The features extracted from these two streams can  complement each other and enhance the accuracy and stability of the overall algorithm. The final layer fuses the outputs of these two streams along with processed demographic information, predicting the probabilities of adverse outcomes or estimating parameters of interest.

In the subsequent subsections, we will give a detailed description of these structures.

\subsection{Sequence-dependent feature extraction stream}\label{subsec4}

After pre-processing, the complex 3D body model is represented by a one-dimensional sequence of level circumferences. This simplified representation can reduce the redundancy of the raw data and mitigate overfitting given small data size. While a back propagation (BP) neural network can extract abstract features from this body shape data, it alone may overlook latent spatial dependencies between sequence elements. 

\begin{figure}
	\centering
	\includegraphics[width=.7\columnwidth]{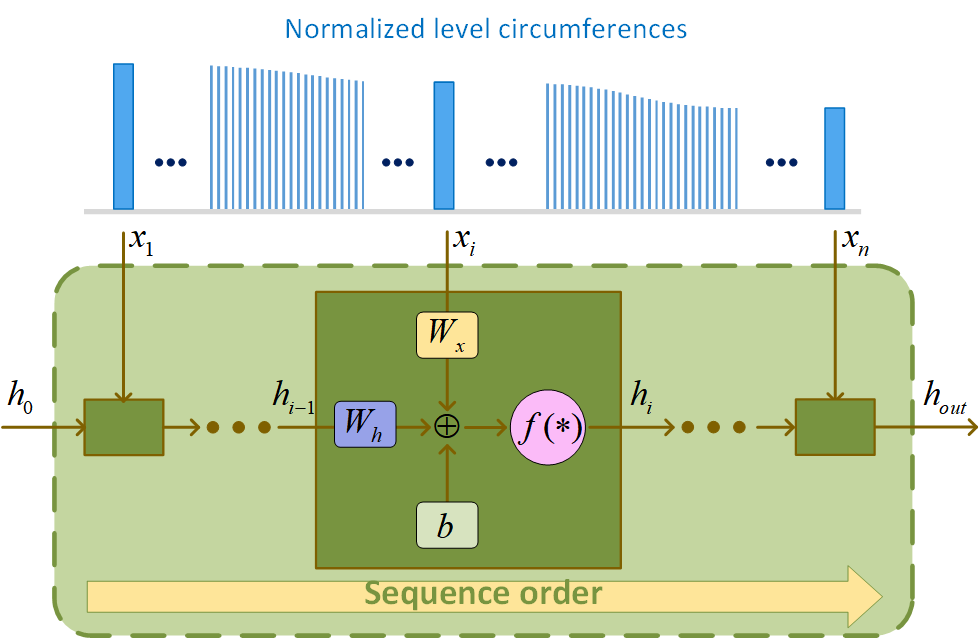}
	\caption{Processing details of the $i$th element.}
	\label{FIG:4}
\end{figure}

Therefore, to maximize the information derived from the level circumference sequence, we implemented an Elman RNN consisting of a single RNN layer. The network contains 64 steps, corresponding to the number of elements in the sequence. The 64 circumference values are sequentially input into the RNN unit based on their sequence index. \autoref{FIG:4} illustrates the calculation at the $i$th time step

Unlike a typical neural network unit, the RNN unit processes not only the current input, but also considers the output from the previous operation. As depicted in \autoref{FIG:4}, the RNN unit processing the element with index $i$ uses the current level circumference value $x_i$ and the previous output $h_{i-1}$ as inputs. The two inputs are given different trainable weights $W_x$ and $W_h$. After adding a trainable bias $b$, the weighted sum is processed through a nonlinear activation function $f(*)$. In our network, we specifically use the hyperbolic tangent function $tanh(*)$ to implement this nonlinear transformation. The output $h_i$ is then sent back to the RNN cell in the next step. The entire computation process can be formulated as follows:
\begin{equation}
h_i=tanh(x_iW^T_x+h_{i-1}W^T_h+b).
\end{equation}
This output is subsequently used in the next step, forming a recurrent operation. This operation facilitates the extraction of local sequential information and its conveyance towards the final output. The output of the last step $h_{out}$ will serve as the body shape feature with spatial information, encapsulating both sequence element values and spatial dependency.

\subsection{Global feature extraction stream}\label{subsec4}

In the global feature extraction stream, we ignore the spatial relationship among level circumferences, with focus instead placed on extracting the most representative features based on their values. Unsupervised learning algorithms are usually used to discover the hidden patterns of data without label guidance. They can explore the data based on its own distribution. In our method, we employ PCA to extract the features for downstream tasks. This algorithm identifies directions in high dimensional space which explain a maximum amount of variance. In other words, it reduces the feature dimensionality while retaining most of the information. 

Since smaller dataset with higher data dimensionality can increase the risk of overfitting, we use PCA to transform the redundant 64-dimensional level circumferences to lower-dimensional global features. By calculating the cumulative explained variance, we discovered that the first three principal components account for 98.1\% of the variance.

Consequently, we use this 3-dimensional feature, $vec_{PCA}$, in subsequent processes as the global feature of the 3D body shape.

\subsection{Joint processing layer}\label{subsec4}

In addition to the 3D body shape data, demographic features also provide crucial information for the prediction and estimation tasks. In our method, we incorporate height, weight and GA as input. Our final joint processing layer concatenates the two body shape feature vectors - $h_{out}$ derived from the sequence-dependent feature extraction stream, and $vec_{PCA}$ from the global feature extraction stream - with the demographic information vector $vec_{basic}$, to make up the fused feature $vec_{fused}$. 

To limit the elements from $h_{out}$ to be at the same scale as the elements from the other two normalized vectors, we used batch normalization operation before the concatenation and obtain the $vec_{RNN}$.
\begin{equation}
vec_{fused}=concat(vec_{RNN},vec_{PCA},vec_{basic}).
\end{equation}
A linear fully connected layer is then applied to generate the final output. The implementation of this operation is represented by the following formula, where $W_L$ and $b_L$ are trainable network parameters:
\begin{equation}
output=vec_{fused}W^T_L +b_L.
\end{equation}
Notice here both the dimensions of $vec_{PCA}$ and $vec_{basic}$ are 3. While processing concatenated features with fully connected layer, the dimension of each feature can also be considered as a “weight” \cite{cheng2023ffa}. Thus, we set the dimension of $h_{out}$ as 5, which is closer to the lower dimension of other features but still able to contain enough information, so that none of these three features can dominantly contribute to the output.
In our experiments, we make minor adjustment on the final layer for different tasks. For regression tasks, we directly use this $output$ value as the final result and use the Mean Square Error (MSE) loss as criterion. For binary classification tasks, we generate a probability ranging from 0 to1 by using sigmoid function on the $output$ and use corresponding Binary Cross-Entropy (BCE) loss as criterion.

\section{Experiments}\label{sec5}

We conducted binary classification tasks using 3D body shape features for the prediction of various adverse pregnancy outcomes, including the risk of preterm labor, GDM, PEC, and the likelihood of undergoing a cesarean section during delivery. We also conducted regression tasks to estimate current fetal weight and MVP with the same inputs. The data were split for cross-validation and independent test set in a 7:3 ratio. A 5-fold cross-validation with early stopping was employed for model tuning to mitigate overfitting and enhance the generalizability of the algorithms.

To assess the effectiveness of employing 3D body shape in these tasks and evaluate the performances of different algorithms, we implemented different evaluation metrics for classification tasks and regression tasks. Heatmaps were used to interpret which body parts contributed more to the classification results, and the Bland-Altman plots were employed to compare the regression performances \cite{altman1983measurement}. Additionally, we designed an ablation study to validate the necessity and impact of critical components of our algorithm.

\subsection{Evaluation metrics}\label{subsec5}

In evaluating the binary classification performance of algorithms, we use metrics including accuracy, precision (also referred to as Positive Predictive Value, PPV), recall, specificity, F1 score, and area under the receiver operating characteristic curve (AUC-ROC). For AUC-ROC score, the Mann-Whitney U-test is employed as a statistical test. With this method, we also calculate the p-value of the AUC-ROC score to show the effectiveness of the algorithms. 

To evaluate the regression performance of algorithms, we use Mean Absolute Error (MAE), Root Mean Squared Error (RMSE), Mean Absolute Percent Error (MAPE, also referred to as Mean Relative Error, MRE), and Root Mean Squared Percent Error (RMSPE) as the evaluation metrics. 
\begin{equation}
MAE=\frac{1}{n}\sum_{n=1}^{n}|\hat{y}_i-y_i|.
\end{equation}
\begin{equation}
RMSE=\sqrt{\frac{1}{n}\sum_{n=1}^{n}(\hat{y}_i-y_i)^2}.
\end{equation}
\begin{equation}
MAPE=\frac{1}{n}\sum_{n=1}^{n}|\frac{\hat{y}_i-y_i}{y_i}|\times100\%.
\end{equation}
\begin{equation}
RMSPE=\sqrt{\frac{1}{n}\sum_{n=1}^{n}(\frac{\hat{y}_i-y_i}{y_i})^2\times100\%}.
\end{equation}
Additionally, we introduce an accuracy metrics which gives the percentage of estimated values falling within an acceptable range. The accuracy metric is defined as follows:
\begin{equation}
Acc=\frac{1}{n}\sum_{n=1}^{n}bool(|\frac{\hat{y}_i-y_i}{y_i}|<m).
\end{equation}
Here, $m$ represents an error tolerance range. We set $m=10\%$ and $m=5\%$ respectively, which is considered acceptable for clinicians \cite{milner2018accuracy}. Corresponding results were calculated  in our experiments.

\subsection{Classification Performance}\label{subsec5}

Binary classification experiments were conducted to predict the risks of preterm labor, GDM, PEC, and the likelihood of undergoing a cesarean section. To evaluate the effectiveness of using 3D body scans compared to traditional anthropometric measurements, we developed a baseline method using anthropometric measurements automatically generated by the optical scanner (height, weight, BMI, waist circumference, hip circumference, waist-to-hip ratio, waist-to-height ratio) and GA as inputs. Logistic Regression (LoR) was used as the baseline model to process these inputs. We also conducted comparison experiments with several popular machine learning algorithms using the same 3D body shape and shape-related demographic features as our algorithm used. These algorithms include LoR, Backpropagation Neural Network (BPNN), Random Forest (RF), and Support Vector Classifier (SVC). Since these algorithms cannot effectively process high dimensional features given small data size, we incorporated PCA to help improve their performance and mitigate potential overfitting problem. 

\begin{table}[h]
\caption{Performance of classification methods for delivery type prediction}\label{tbl3}
\begin{tabular*}{\textwidth}{@{}l @{\extracolsep{\fill}} lllllll @{}}
\toprule
 & Accuracy & Precision & Recall & Specificity & F1 Score & AUC-ROC (p-value) \\
\midrule
Baseline & 74.36\% & 50.00\% & 30.00\% & 89.66\% & 0.38 & 0.590 (0.4012) \\
PCA+LoR & 71.79\% & 42.86\% & 30.00\% & 86.21\%	& 0.35 & $<0.5$ (N/A)\\
PCA+BPNN & 69.23\% & 41.67\% & \bfseries 50.00\% & 75.86\% & \bfseries 0.45 & 0.638 (0.1980) \\
PCA+RF & 53.85\% & 27.78\% & \bfseries 50.00\% & 55.17\% & 0.36 & 0.534 (0.7511) \\
PCA+SVC & 64.10\% & 33.33\% & 40.00\% & 72.41\% & 0.36 & 0.566 (0.5382) \\
Ours & \bfseries 79.49\% & \bfseries 75.00\% & 30.00\% & \bfseries 96.55\% & 0.43 & \bfseries 0.641 (0.1885) \\
\bottomrule
\end{tabular*}
\begin{minipage}{\textwidth}
\vspace{0.1cm}
\footnotesize Notes: The p-value in parentheses is calculated using the Mann-Whitney U test on the AUC-ROC.
\end{minipage}

\caption{Performance of classification methods for preterm labor prediction}\label{tbl4}
\begin{tabular*}{\textwidth}{@{}l @{\extracolsep{\fill}} lllllll @{}}
\toprule
 & Accuracy & Precision & Recall & Specificity & F1 Score & AUC-ROC (p-value) \\
\midrule
Baseline & 79.49\% & 28.57\% & 40.00\% & 85.29\% & 0.33 & \bfseries 0.771 (0.0530) \\
PCA+LoR & \bfseries 89.74\% & \bfseries 66.67\% & 40.00\% & \bfseries 97.06\% & \bfseries 0.50 & 0.588 (0.5297) \\
PCA+BPNN & 87.18\% & 50.00\% & 20.00\% & \bfseries 97.06\% & 0.29 & 0.659 (0.2562) \\
PCA+RF & 69.23\% & 23.08\% & \bfseries 60.00\% & 70.59\% & 0.33 & 0.647 (0.2938) \\
PCA+SVC & 46.15\% & 13.64\% & \bfseries 60.00\% & 44.12\% & 0.22 & 0.582 (0.5581) \\
Ours & \bfseries 89.74\% & \bfseries 66.67\% & 40.00\% & \bfseries 97.06\% & \bfseries 0.50 & 0.676 (0.2088) \\
\bottomrule
\end{tabular*}
\begin{minipage}{\textwidth}
\vspace{0.1cm}
\footnotesize Notes: The p-value in parentheses is calculated using the Mann-Whitney U test on the AUC-ROC.
\end{minipage}

\caption{Performance of classification methods for GDM prediction}\label{tbl5}
\begin{tabular*}{\textwidth}{@{}l @{\extracolsep{\fill}} lllllll @{}}
\toprule
 & Accuracy & Precision & Recall & Specificity & F1 Score & AUC-ROC (p-value) \\
\midrule
Baseline & 85.37\% & 40.00\% & 40.00\% & 91.67\% & 0.40 & 0.667 (0.2311) \\
PCA+LoR & 85.37\% & 40.00\% & 40.00\% & 91.67\% & 0.40 & 0.683 (0.1894) \\
PCA+BPNN & 87.80\% & 50.00\% & 20.00\% & \bfseries 97.22\% & 0.29 & 0.600 (0.4733) \\
PCA+RF & 87.80\% & 50.00\% & 20.00\% & \bfseries 97.22\% & 0.29 & 0.556 (0.6880) \\
PCA+SVC & 85.37\% & 40.00\% & 40.00\% & 91.67\% & 0.40 & 0.694 (0.1642) \\
Ours & \bfseries 92.68\% & \bfseries 75.00\% & \bfseries 60.00\% & \bfseries 97.22\% & \bfseries 0.67 & \bfseries 0.800 (\bfseries 0.0314) \\
\bottomrule
\end{tabular*}
\begin{minipage}{\textwidth}
\vspace{0.1cm}
\footnotesize Notes: The p-value in parentheses is calculated using the Mann-Whitney U test on the AUC-ROC.
\end{minipage}

\caption{Performance of classification methods for PEC prediction}\label{tbl6}
\begin{tabular*}{\textwidth}{@{}l @{\extracolsep{\fill}} lllllll @{}}
\toprule
 & Accuracy & Precision & Recall & Specificity & F1 Score & AUC-ROC (p-value) \\
\midrule
Baseline & 66.67\% & 15.38\% & \bfseries 50.00\% & 68.57\% & 0.24 & 0.529 (0.8509) \\
PCA+LoR & 76.92\% & 14.29\% & 25.00\% & 82.86\% & 0.18 & 0.557 (0.7118) \\
PCA+BPNN & 84.62\% & 25.00\% & 25.00\% & 91.43\% & 0.25 & 0.586 (0.5773) \\
PCA+RF & \bfseries 89.74\% & \bfseries 50.00\% & 25.00\% & \bfseries 97.14\% & \bfseries 0.33 & \bfseries 0.632 (0.3923) \\
PCA+SVC & 79.49\% & 16.67\% & 25.00\% & 85.71\% & 0.20 & 0.500 (1.0000) \\
Ours & 84.62\% & 25.00\% & 25.00\% & 91.43\% & 0.25 & 0.625 (0.4179) \\
\bottomrule
\end{tabular*}
\begin{minipage}{\textwidth}
\vspace{0.1cm}
\footnotesize Notes: The p-value in parentheses is calculated using the Mann-Whitney U test on the AUC-ROC.
\end{minipage}
\end{table}

The results on the test sets are shown in \autoref{tbl3}, \ref{tbl4}, \ref{tbl5}, \ref{tbl6}, with the best performances highlighted in bold. We considered the situations with adverse outcomes as the positive class. For instance, in the delivery type classification task, participants who underwent a cesarean section are labeled as positive.

These results suggest that 3D body shapes captured in the second trimester can be effectively used to predict the risk of preterm labor and GDM, achieving accuracy rates exceeding 89\%. However, its performance is not very good in predicting the delivery type and PEC, where all of the methods get an F1 score lower than 0.5 and AUC-ROC lower than 0.650. This may be attributed to the complex factors influencing delivery type and the risk of PEC, such as obstetric history and gestational disease history, which were not considered in this study.

By comparing the performances of the baseline method (LoR using anthropometric measurements as input) with LoR using 3D body shape data for preterm labor and GDM prediction, we find that the same LoR model achieves better performance when using 3D body shapes rather than traditional anthropometric measurements. Compared with other algorithms, our proposed method shows superior performance in predicting the risk of preterm labor and GDM. Especially for GDM prediction, our algorithm outperforms all others across all metrics, with a p-value below 0.05. Even in the more challenging delivery type prediction task, our algorithm achieves the highest accuracy, precision, specificity, and AUC-ROC, although BPNN and RF achieve a higher recall score of 50.00\%. These results suggest that 3D body shape data have greater potential than anthropometric measurements for predicting adverse pregnancy outcomes and validate the effectiveness of our approach in extracting useful information from 3D body scans.

\begin{figure*}
	\centering
	\includegraphics[width=\textwidth]{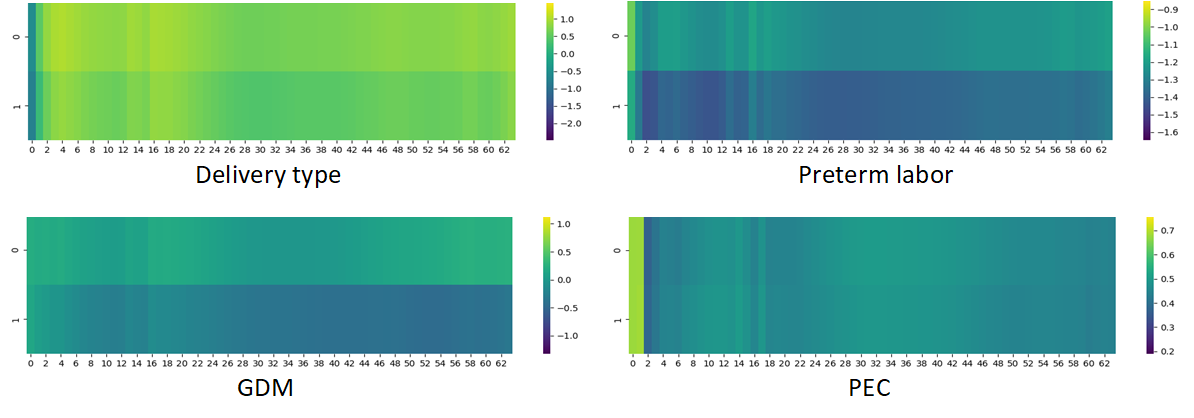}
	\caption{Heatmaps of average RNN hidden states when processing the 64 level circumferences for different tasks. The upper rows labeled "0" correspond to negative-class samples, and the lower rows labeled "1" correspond to positive-class samples.}
	\label{FIG:5}
\end{figure*}

To provide a more intuitive understanding of which regions of abdominal body contribute more to predict adverse pregnancy outcomes, \autoref{FIG:5} shows the average hidden state values for the negative and positive classes separately. From left to right, the 64 time points correspond to 64 level circumferences measured from the lowest point of the pubic bone to the bottom of the breasts. Referring to \autoref{FIG:2}, level circumferences 1–20 approximately cover the hip region, and level circumferences 30–40 roughly cover the waist region. For all tasks, the lower hip, top of the pelvis, and upper abdominal regions appear to play the most important roles, as indicated by greater changes in the hidden states. Except in the GDM prediction task, our model assigns extra attention to the lower hip region. Furthermore, we observe that for tasks on which our model performs well, there are clear differences between the heatmaps of negative and positive samples. In contrast, for tasks with poorer performance, the heatmaps are more similar, suggesting that the RNN branch may not extract sufficiently informative features from the body shape for downstream classifications.

\subsection{Regression Performance}\label{subsec5}

In addition to predicting the risk of adverse outcomes, we also explored the use of 3D body shape for estimating current intrauterine parameters. We adjusted the final layer of our network and conducted regression experiments to estimate fetal weight and MVP. A baseline method was designed using a Linear Regression (LR) model with traditional anthropometric measurements as inputs, and we also included the mean of the ground truth values as an additional baseline. LR, BPNN, RF, and SVR with PCA were evaluated using the same inputs as in the classification tasks for comparison. The results on the test sets are presented in \autoref{tbl7} and \autoref{tbl8} below. The best results are also highlighted in bold.

\begin{table}[h]
\caption{Performance of models for EFW estimation}\label{tbl7}
\begin{tabular*}{\textwidth}{@{}l @{\extracolsep{\fill}} lllllll @{}}
\toprule
 & MAE(g) & RMSE(g) & MAPE & RMSPE & Acc(m=10\%) & Acc(m=5\%) \\
\midrule
Baseline & 34.80 & 44.53 & 9.43\% & 11.57\% & 61.11\% & 30.56\% \\
Average	& 43.53 & 62.03 & 11.20\% & 14.53\% & 55.56\% & 30.56\% \\
PCA+LR & 39.25 & 51.88 & 10.51\% & 13.63\% & 58.33\% & 30.56\% \\
PCA+BPNN & 36.03 & 50.24 & 9.62\% & 13.03\% & 61.11\% & 38.89\% \\
PCA+RF & 33.54 & 45.09 & 9.08\% & 11.84\% & 61.11\% & 41.67\% \\
PCA+SVR & 34.12 & 44.37 & 9.07\% & 11.39\% & 61.11\% & 30.56\% \\
Ours & \bfseries 30.40 & \bfseries 42.78 & \bfseries 8.19\% & \bfseries 10.94\% & \bfseries 72.22\% & \bfseries 44.44\% \\
\bottomrule
\end{tabular*}

\caption{Performance of models for MVP estimation}\label{tbl8}
\begin{tabular*}{\textwidth}{@{}l @{\extracolsep{\fill}} lllllll @{}}
\toprule
 & MAE(cm) & RMSE(cm) & MAPE & RMSPE & Acc(m=10\%) & Acc(m=5\%) \\
\midrule
Baseline & 0.718 & 0.880 & 16.80\% & 21.69\% & 35.29\% & 23.53\% \\
Average	& \bfseries 0.542 & \bfseries 0.727 & \bfseries 12.21\% & \bfseries 16.22\% & \bfseries 50.00\% & \bfseries 41.18\% \\
PCA+LR & 0.921 & 1.075 & 21.20\% & 25.72\% & 26.47\% & 11.76\% \\
PCA+BPNN & 0.696 & 0.873 & 16.19\% & 21.28\% & 44.12\% & 20.59\% \\
PCA+RF & 0.672 & 0.848 & 15.57\% & 20.35\% & 35.29\% & 29.41\% \\
PCA+SVR & 0.672 & 0.851 & 15.80\% & 20.67\% & \bfseries 50.00\% & 20.59\% \\
Ours & 0.625 & 0.801 & 14.50\% & 19.21\% & \bfseries 50.00\% & 29.41\% \\
\bottomrule
\end{tabular*}

\end{table}

\begin{figure}
	\centering
	\includegraphics[width=\textwidth]{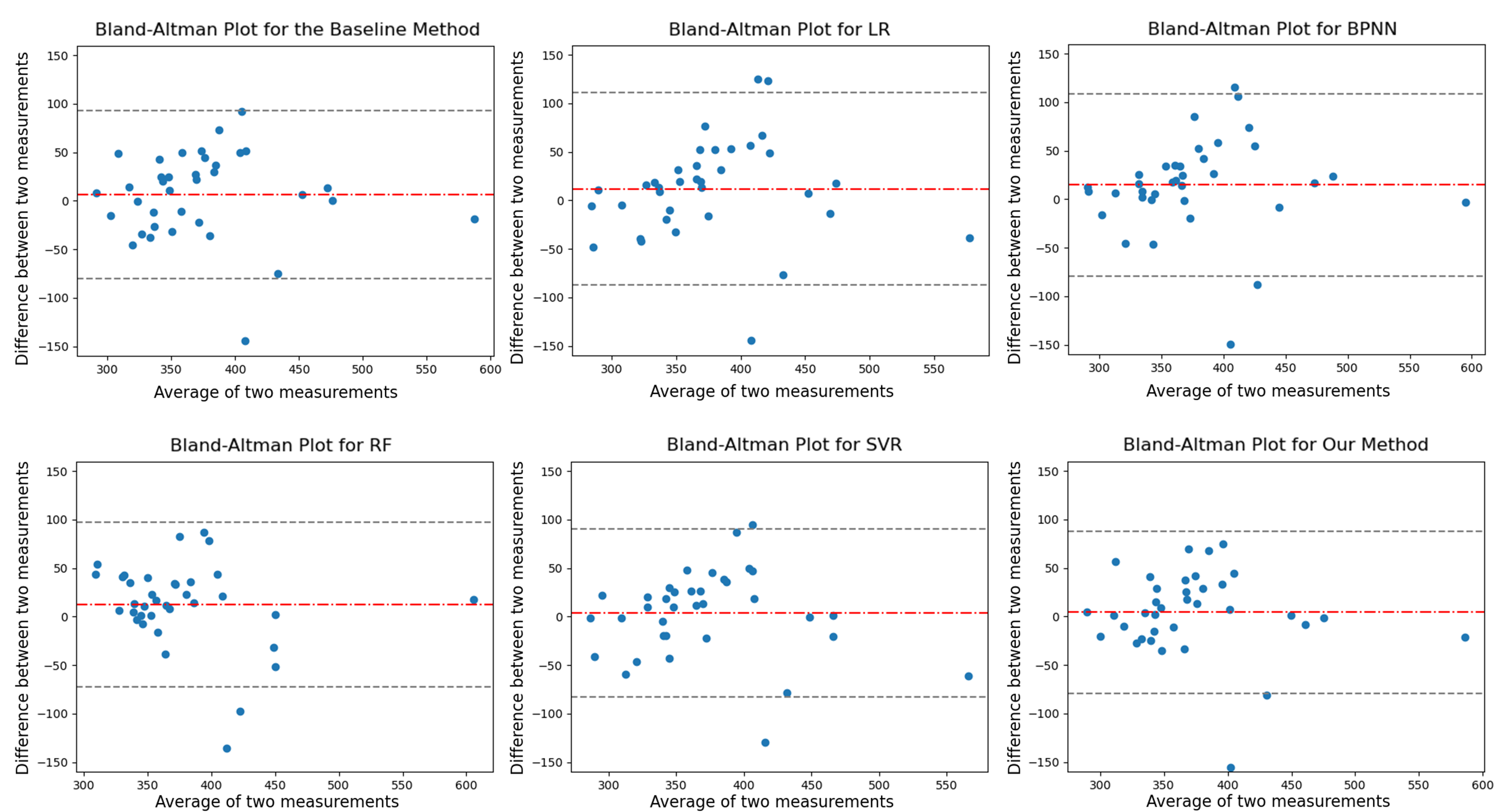}
	\caption{Bland-Altman plot of regression results for EFW.}
	\label{FIG:6}
\end{figure}

The results in \autoref{tbl7} indicate that our algorithm outperforms other algorithms in estimating EFW across all metrics. While the SVR model also demonstrates commendable performance, our algorithm exceeds it by 10.90\% in MAE, 3.58\% in RMSE, 9.70\% in MAPE, 3.95\% in RMSPE, and achieves an accuracy of 72.22\% within a 10\% error tolerance. Interestingly, the baseline method also shows promising results, outperforming the same LR model using 3D body shape as inputs. \autoref{FIG:6} illustrates the estimation results for each participant using different algorithms with Bland-Altman plots, which are widely used to analyze the agreement between two clinical measurements in biomedicine. In these plots, the mean of the differences is shown as a red dash-dotted line, and the 95\% limits of agreement are represented by two dashed lines. From \autoref{FIG:6}, we find that the estimations of our method distribute more tightly within a narrower agreement interval. This finding aligns with the observations from \autoref{tbl7}, demonstrating the utility of 3D body shape in estimating fetal weight. We also investigated the mutual outlier with an average value around 410 and find that its EFW percentile among all fetuses at the same GA is 99\%. It indicates that body shape–based methods still have room for improvement in extreme cases. Moreover, by observing the potential fitted line of the scatter plots, we find that all of these methods tend to underestimate the lower fetal weights and overestimate the higher fetal weights, which can be addressed in future work.

\begin{figure}
	\centering
	\includegraphics[width=\textwidth]{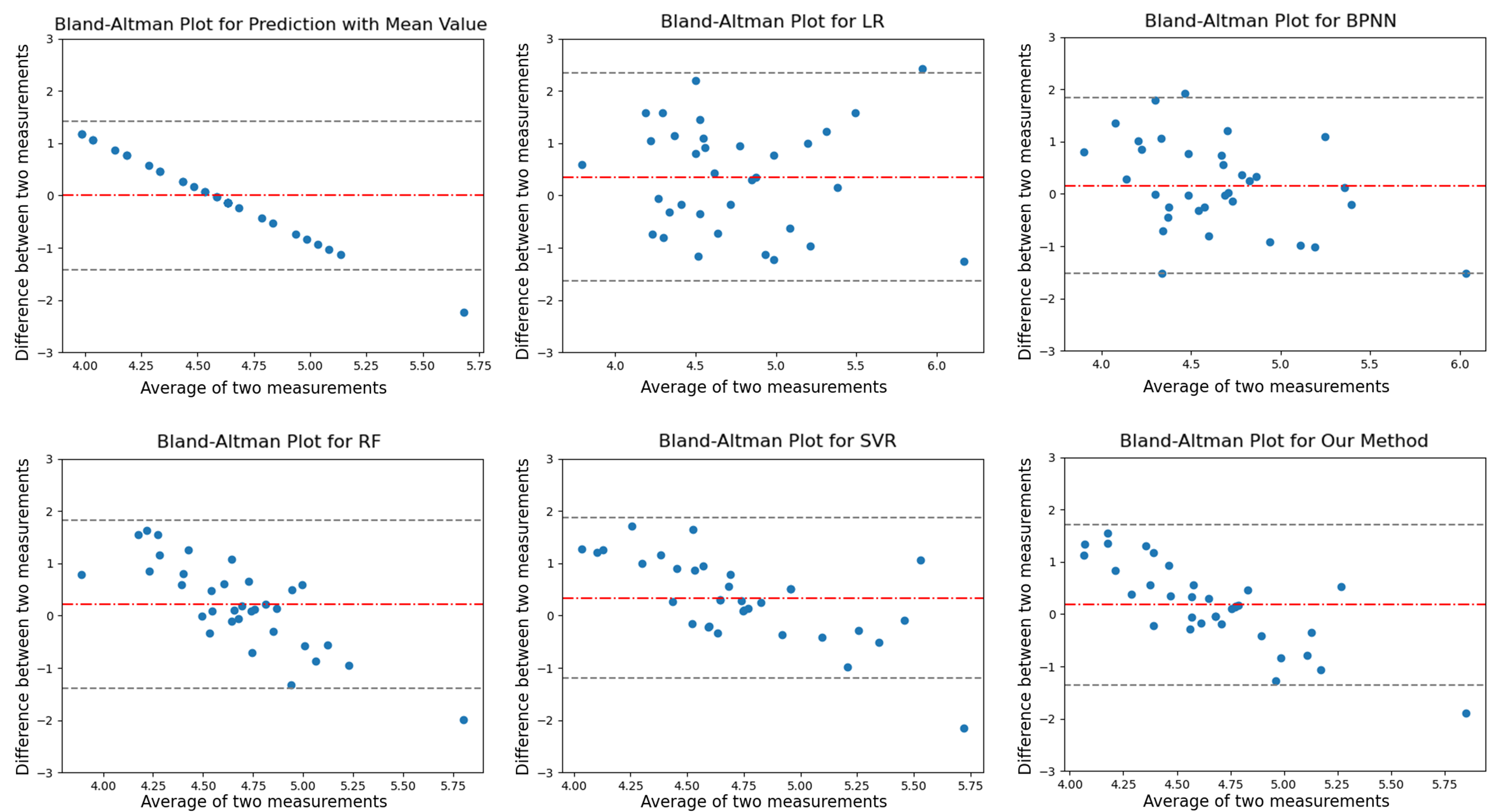}
	\caption{Bland-Altman plot of regression results for MVP.}
	\label{FIG:7}
\end{figure}

When estimating MVP using these algorithms, the performances are not as good as when estimating fetal weights, as shown in \autoref{tbl8}. For machine learning methods other than ours, the MAPEs and RMSPEs on this task exceeded 15\% and 20\% respectively, which is notably higher than the roughly 9\% MAPEs and 13\% RMSPEs for fetal weight estimation. Although our method achieves the highest accuracy, none of the machine learning methods outperform the simple baseline using the mean of the ground truth. This suggests that 3D body shape may not be an effective predictor of MVP. The corresponding Bland-Altman plots are shown in \autoref{FIG:7}. From these plots, we find that none of the machine learning methods achieve a narrower agreement interval than the mean-value prediction. Moreover, based on the fitted lines of the scatter plots, the three methods listed in the bottom row of \autoref{FIG:7} tend to predict values close to the average rather than capturing meaningful patterns. This unexpected result may be due to fluctuations caused by factors such as fetal position and maternal hydration \cite{harman2008amniotic, ulker2013effect}. Additionally, MVP is less related to maternal nutritional status, which is a key feature captured by body shape–based methods.

\subsection{Ablation Study}\label{subsec5}

Our proposed method exhibits overall good performance in both classification and regression tasks, which may be attributed to the separate application of PCA and RNN to extract complementary features. To substantiate this claim, we conducted a set of experiments that estimate fetal weight using networks with or without either of these two streams. Additionally, we explored replacing the RNN with the gated recurrent unit (GRU), an enhanced version of the vanilla RNN, to extract sequential information from level circumferences. The GRU is designed with ability of learning both long-term and short-term dependencies, while having fewer parameters than the long short-term memory (LSTM) network. Through comparisons with this GRU network, we aim to demonstrate that our network architecture performs better with the conventional RNN. \autoref{tbl9} lists the results of the various combinations we tested during the algorithm development process with the best
performances highlighted in bold.

\begin{table}[h]
\caption{Performances on estimating fetal weight with different network architectures. The "FC" denotes the fully connected layer.}\label{tbl9}
\begin{tabular*}{0.85\textwidth}{@{}l @{\extracolsep{\fill}} llllll @{}}
\toprule
 & MAE & RMSE & MAPE & RMSPE & Acc(m=10\%) \\
\midrule
FC & 37.09 & \bfseries 42.18 & 9.98\% & 11.37\% & 52.78\% \\
PCA+FC & 36.03 & 50.24 & 9.62\% & 13.03\% & 61.11\% \\
RNN+FC & 34.05 & 51.50 & 8.68\% & 11.68\% & \bfseries 72.22\% \\
PCA+GRU+FC & 34.11 & 44.60 & 9.23\% & 11.57\% & 61.11\% \\
Ours\\(PCA+RNN+FC)  & \bfseries 30.40 & 42.78 & \bfseries 8.19\% & \bfseries 10.94\% & \bfseries 72.22\% \\
\bottomrule
\end{tabular*}
\end{table}

Comparing the results between the second and third rows in \autoref{tbl9}, we find that PCA facilitates FC in learning more useful information from raw level circumferences, leading to a 1.03g decrease in MAE and 15.78\% increase in accuracy. Further comparison between the third, fourth and last rows indicates that using either PCA or RNN alone does not yield optimal results. The incorporation of both, as implemented in our final design, presents the most significant improvement in performance. The fifth row shows the performance results when the RNN is replaced by GRU in our network. Despite the GRU's advanced capabilities in capturing long-term relationships from sequences, this modification does not bring more improvement in this task. This failure could potentially be due to the fact that GRU contains more trainable parameters than the conventional RNN, thus resulting in reduced learning efficiency, particularly when dealing with small data size. These findings affirm the effectiveness of our proposed hybrid structure for 3D body feature extraction.

\section{Conclusion}\label{sec6}

In this study, we investigated the potential of using 3D body scans with basic body shape-related demographic information for prenatal care in future telehealth applications and explored how to use this modality effectively. We proposed a novel neural network which incorporates a supervised learning stream and an unsupervised learning stream for extracting features from sampled abdominal level circumferences. The supervised stream leverages RNN units to extract sequential information from level circumferences, while the unsupervised stream employs the PCA to captures global descriptors of abdominal body shape. We applied this algorithm to predict the risk of preterm labor, GDM, PEC, the likelihood of cesarean section, and to estimate current fetal weight and MVP. The results indicate that maternal 3D body shape, captured during 18-24 gestational weeks, is effective in predicting GDM and preterm labor, as well as estimating fetal weight. Compared to other well-performing machine learning algorithms, our proposed method demonstrates superior performances in these tasks.

Nevertheless, there is room for future improvement. Firstly, our current dataset is small and only contains scans obtained in patients’ second trimester of pregnancy, potentially limiting the model's generalizability to estimate fetal weight in other trimesters, since fetal growth trends may vary across different stages. Ideally, with the inclusion of actual birth weights and 3D body scans closer to delivery, a more accurate model may be developed guided by neonatal weights. We may also explore the potential of using longitudinal data for better estimation. Secondly, our current analysis focused solely on the abdominal region, overlooking other body parts. Given the research \cite{boucher2022maternal,tosson2023neonatal,li2018first,mazhar2018correlation} suggesting correlations between appendicular body shape and pregnancy outcomes, future studies could investigate methods using the entire body shape to get more accurate estimations and predictions. Thirdly, this pilot study only explored the usability of body shape for the discussed tasks, however, additional easily accessible medical variables, such as obstetric history and chronic disease history, could be incorporated to increase model precision. With these variables incorporated, it can be more practically applicable for intelligent, remote prenatal care. Additionally, to further facilitate the application of 3D body scans in telehealth, we will also conduct extended experiments using 3D models obtained from smartphone scans.

\bmhead{Conflict of interest}

The authors have no conflicts of interest to disclose.

\bmhead{Data availability statement}

Data may be made available based on the nature of the request.

\bmhead{Funding statement}

Research reported in this publication was supported by the National Institute of Diabetes and Digestive and Kidney Diseases of the National Institutes of Health under Award Number R01DK129809. The content is solely the responsibility of the authors and does not necessarily represent the official views of the National Institutes of Health.

\input{my-paper.bbl}

\section*{Author Biographies}

\textbf{Ruting Cheng} is a Ph.D. student at George Washington University. She received her M.S. degree from the University of Science and Technology Beijing in 2022. Her research interests include computer vision and machine learning in medicine.

\textbf{Yijiang Zheng} is a Ph.D. student at George Washington University, where he earned his master’s degree in computer science in 2020. His research focuses on 3D computer vision applications within the medical field.

\textbf{Boyuan Feng} is a Ph.D. student in Computer Science at George Washington University and holds a B.S. from Huazhong University of Science and Technology. Her research focuses on machine learning and computer vision.

\textbf{Chuhui Qiu} is a graduate student at the George Washington University, where he received his M.S. degree. His field of study is machine learning, computer vision, and algorithm design.

\textbf{Zhuoxin Long} is currently a Ph.D. student at the George Washington University, where she earned her master's degree in Biostatistics. Her research interests include causal inference and individualized treatment regimes.

\textbf{Joaquin A. Calderon} earned his MD from Universidad del Norte in Barranquilla, Colombia in 2018. He joined the GW Medical Faculty Associates in 2022 as a Research Coordinator in Obstetrics and Gynecology. He is interested in maternal-fetal medicine.

\textbf{Xiaoke Zhang} is currently an Associate Professor in the Department of Statistics at George Washington University. His research interests include longitudinal data analysis, statistical learning, causal inference and applied statistics.

\textbf{Jaclyn M. Phillips}, MD is a Maternal Fetal Medicine specialist and Assistant Professor of Obstetrics \& Gynecology. Her research interests include postpartum hemorrhage, severe maternal morbidity, peripartum blood management, and risk prediction.

\textbf{James K. Hahn} is a Professor in both the Department of Computer Science, and School of Medicine and Health Sciences at the George Washington University. His areas of interests are: medical simulation, machine learning and medical visualization.

\end{document}